\documentclass[conference]{IEEEtran}
\IEEEoverridecommandlockouts
\usepackage{cite}
\usepackage{amsmath,amssymb,amsfonts}
\usepackage{algorithmic}
\usepackage{graphicx}
\usepackage{textcomp}
\usepackage{xcolor}
\usepackage{url}

\usepackage{array}
\usepackage{color, colortbl}
\usepackage{booktabs}
\usepackage{makecell, bm} 
\usepackage{latexsym}
\usepackage[ruled,linesnumbered]{algorithm2e}
\def\BibTeX{{\rm B\kern-.05em{\sc i\kern-.025em b}\kern-.08em
    T\kern-.1667em\lower.7ex\hbox{E}\kern-.125emX}}
\begin{document}

\def\BibTeX{{\rm B\kern-.05em{\sc i\kern-.025em b}\kern-.08em
    T\kern-.1667em\lower.7ex\hbox{E}\kern-.125emX}}
\def\trianPDF#1#2{0 0 m #1 0 l #2 4.5 l h}
\def\uptrianPDF#1#2{#2 0 m #1 4.5 l 0 4.5 l h}

\def\credCOLOR   {.54 .14 0}
\def\cblueCOLOR  {.16 .8 .94}
\def\cgreenCOLOR {0.38, 1, 0.2}

\def\genbox#1#2#3#4#5#6{
\leavevmode\raise#4bp\hbox to#5bp{\vrule height#5bp depth0bp width0bp
\pdfliteral{q .5 w \csname #2COLOR\endcsname\space RG
                   \csname #3PDF\endcsname{#5}{#6} S Q
         \ifx1#1 q \csname #2COLOR\endcsname\space rg 
                   \csname #3PDF\endcsname{#5}{#6} f Q\fi}\hss}}

\def\trianbox   #1#2{\genbox{#1}{#2}  {trian}    {0}   {5}    {2.5}}
\def\uptrianbox #1#2{\genbox{#1}{#2}  {uptrian}  {0}   {5}    {2.5}}

\title{Mind2: Mind-to-Mind Emotional Support System with Bidirectional Cognitive Discourse Analysis}

\author{
\IEEEauthorblockN{Shi Yin Hong, Uttamasha Oyshi, Quan Mai, Gibson Nkhata, and Susan Gauch}
\IEEEauthorblockA{\textit{Department of Electrical Engineering and Computer Science} \\
\textit{University of Arkansas} \\
Fayetteville, Arkansas, USA  \\
\{syhong, uoyshi, quanmai, gnkhata, sgauch\}@uark.edu}
}

\maketitle

\begin{abstract}
Emotional support (ES) systems alleviate users' mental distress by generating strategic supportive dialogues based on diverse user situations. 
However, ES systems are limited in their ability to generate effective ES dialogues that include timely context and interpretability, hindering them from earning public trust. 
Driven by cognitive models, we propose Mind-to-Mind (Mind2), an ES framework that approaches interpretable ES context modeling for the ES dialogue generation task from a discourse analysis perspective. 
Specifically, we perform cognitive discourse analysis on ES dialogues according to our dynamic discourse context propagation window, which accommodates evolving context as the conversation between the ES system and user progresses.
To enhance interpretability, Mind2 prioritizes details that reflect each speaker's belief about the other speaker with bidirectionality, integrating Theory-of-Mind, physiological expected utility, and cognitive rationality to extract cognitive knowledge from ES conversations. Experimental results support that Mind2 achieves competitive performance versus state-of-the-art ES systems while trained with only 10\% of the available training data. 


\end{abstract}


\section{Introduction}

Fueled by recent developments in large language models (LLMs), advancements in automated supportive dialogue generation performance of emotional support (ES) systems unleash possibilities for them to serve as emotional support assistance in everyday life \cite{song2024typing}. Not only do ES systems serve as accessible tools to provide service for those in need to overcome barriers to acquire timely support, but they also possess the potential to collaborate with psychiatric professionals to improve the efficiency in helping more patients amid professional shortages in related domains \cite{kaziunas2019precarious}\cite{pendse2023marginalization}. While demands for live ES systems are growing, adoptions of ES systems in real-world applications encounter uncertainties \cite{ernala2022reintegration}. A main factor in this resistance relates to the limited accountability of ES systems to generate adaptive strategic dialogues with contextual interpretability.

The contextual modeling capability of ES systems contributes to their dialogue generation performance. Effective ES systems generate contextual dialogues according to the progressing needs of the user. To supply context in ES systems, one mainstream approach focuses on users' emotions. These works predominantly utilize user emotions from user dialogues as a source for understanding users' mental states in designing models \cite{lin-etal-2019-moel} \cite{li-etal-2020-empdg} \cite{majumder-etal-2020-mime} \cite{Li_Li_Ren_Ren_Chen_2022} \cite{zheng-etal-2021-comae}. In turn, the context modeling of these ES systems depends on the accuracy of the emotion recognition of dialogues in categorizing user emotions based on supplied features.
Moreover, given the complex human emotional boundaries, utilizing user emotions in categorical representations cannot guarantee optimal user emotion analysis \cite{9330790}. Another branch of ES systems adopts a retrieval-based method, incorporating external knowledge bases to enhance context understanding of target ES scenario with commonsense knowledge \cite{zhou-etal-2023-case} \cite{cai-etal-2023-improving} \cite{tu-etal-2022-misc} \cite{xu2024dynamic}. The assumption is that external knowledge bases contain similar past situations confronted by diverse users. Depending on the choice of external knowledge bases, no optimal commonsense knowledge may be found, or the same commonsense knowledge may be utilized by multiple ES scenarios of users confronting similar situations. In these cases, extracted contexts possess limited situational and temporal relevance.

In this work, we propose Mind-to-Mind (Mind2), an ES framework that tackles context modeling for ES dialogue generation with explicable contextual interpretability. We incorporate bidirectional cognitive discourse analysis: While previous works approach context modeling tailored to the states of users, we consider the perspectives of both the users and the systems. To promote the situational relevancy of Mind2's context modeling, we utilize individual ES conversations as the knowledge source instead of accessing external knowledge. We perform bidirectional cognitive discourse analysis based on the evolving discourse context propagation window to control the local discourse span, supporting the temporal adaptability of Mind2. We employ conceptual foundations from Theory-of-Mind, psychological expected utility, and cognitive rationality to drive prompt-based query expansion during bidirectional cognitive discourse analysis. 
Finally, we utilize the three resulting components of bidirectional cognitive knowledge for ES dialogue generation. In summary, our main contributions are:

\begin{itemize}

    \item We present Mind2, an ES framework that encapsulates cognitive theories to support context modeling interpretability in bidirectional cognitive discourse analysis for ES dialogue generation.

    \item Experimental results show that Mind2 achieves competitive performance against state-of-the-art ES systems, utilizing as little as 10\% training data.

    \item We study the relative effectiveness of three components of bidirectional cognitive knowledge in improving ES dialogue generation of Mind2.

\end{itemize}

\section{Related Work}
\subsection{Emotion Analysis and Emotional Support (ES) Systems} In the ES systems domain, a branch of work focuses on capturing user emotions in understanding ES contexts for ES dialogue generation. MoEL \cite{lin-etal-2019-moel} represents an early model that generates dialogues utilizing emotion distribution and emotion-aware listeners. EmpDG \cite{li-etal-2020-empdg} captures user emotions at token and dialogue scopes with feedback. MIME \cite{majumder-etal-2020-mime} considers users' emotions from a polarity-based perspective with stochasticity. In \cite{Li_Li_Ren_Ren_Chen_2022}, Li et al. propose to generate empathetic responses by constructing emotion context graphs and learning emotional dependencies with an emotional cross-attention mechanism. Bao et al. propose MFF-ESC \cite{bao-etal-2024-multi-stream} to capture emotion intensity transition, fusing text semantics, emotion, and feedback information to tackle the ES task. Similarly, instead of solely focusing on emotion analysis, CoMAE \cite{zheng-etal-2021-comae} integrates the user content from emotions with communication mechanism and dialog act to enhance context understanding.

\subsection{Commonsense Knowledge-based Emotional Support (ES) Systems} Another mainstream method is to adopt commonsense knowledge to enhance context modeling. Commonly, the contextual knowledge is extracted from an external source. 
CEM \cite{Sabour2021CEMCE} further utilizes the ATOMIC \cite{sap2019atomic} as the knowledge base, to integrate user emotion states with cognitive understanding of the user situation. CASE \cite{zhou-etal-2023-case} models emotions and cognition of users jointly as emotional concept and cognition graphs using COMET \cite{bosselut-etal-2019-comet} and ConceptNet \cite{Speer_Chin_Havasi_2017} as sentence-level and word-level knowledge bases, respectively. 
Likewise, in \cite{cai-etal-2023-improving}, Cai et al. approach dialogue generation using commonsense knowledge extracted from COMET \cite{bosselut-etal-2019-comet} but propose methods to probe context with selectivity.
Xu et al. \cite{xu2024dynamic} further integrate persona information with cognitive relationships sourced from the ATOMIC \cite{sap2019atomic} knowledge base. 
In \cite{fu-etal-2023-reasoning}, Fu et al. utilize commonsense knowledge for causality explanation in generating responses. Peng et al. introduce GLHG \cite{Peng2022ControlGU}, which focuses on modeling hierarchical relationships in a global-to-local technique in analyzing the user distress causes and understanding user context integrated with COMET \cite{bosselut-etal-2019-comet}. 
\begin{figure}
    \centering
    \includegraphics[width=1\linewidth]{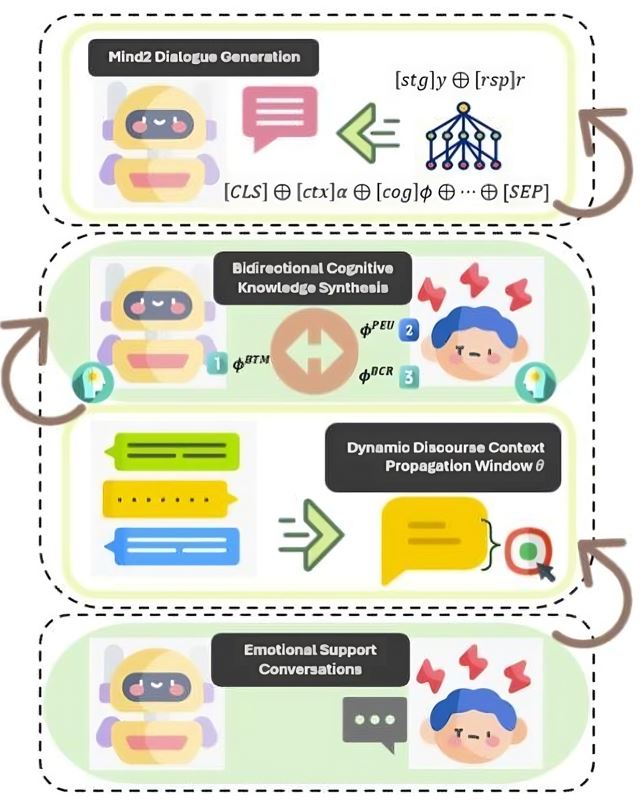}
    \caption{An overview of the Mind2 framework.}
    \label{fig:v2}
    \vspace{-4mm}
\end{figure}

\subsection{Strategic-centric Emotional Support (ES) Systems} Recent works also assess the strategy of support demonstrated in ES dialogue generation. In \cite{liu-etal-2021-towards}, Liu et al. curated the ESConv dataset, which assigns each ES dialogue to one of the eight support strategies. Following, Tu et al. propose MISC \cite{tu-etal-2022-misc} to tackle ES, incorporating ES strategy while considering user emotion features using the COMET \cite{bosselut-etal-2019-comet} knowledge graphs. Peng et al. introduce FADO \cite{fado}, which attends the strategy-to-context flow in ES dialogue history in performing ES dialogues constrained to ES strategies. Cheng et al. consider persona information in ES dialogue generation in their PAL framework \cite{cheng-etal-2023-pal} by analyzing the persona of the support seeker generating ES strategy-constrained dialogues. In MultiESC \cite{cheng-etal-2022-improving}, Cheng et al. focus on optimizing support strategies in a multi-turn ES conversation setting with lookahead heuristics. Similarly, Deng et al. approach ES dialogue generation from a mix-initiative perspective, introducing the KEMI framework \cite{kemi}, but analyze ES dialogues based on their initiative types according to the role of speakers. Likewise, Zhou et al. consider ES from a multi-turn perspective but instead focus on eliciting positive emotion during the support session in their SUPPORTER framework \cite{zhou-etal-2023-facilitating}. 


\section{Methodology}
In this section, we detail the Mind2 framework for ES dialogue generation. Our Mind2 framework tackles the ES context modeling with bidirectional cognitive discourse analysis, stimulating interpretability in extracted contexts with cognitive theories. First, we acquire interpretable bidirectional context with a conceptual basis from the synergy of three cognitive theories based on an evolving discourse context propagation window. Then, we encapsulate the collected contextual knowledge in ES dialogue generation. Figure \ref{fig:v2} illustrates an overview of Mind2.

\subsection{Problem Statement}
Given a collection of $M$ conversations of ES scenarios 
$\mathcal{D} = \{{\mathbf{d}}^{m}, \mathbf{r}^{m}\}_{m = 1}^{M}$, we perform the task of ES dialogue generation in a multi-turn setting. The dialogue history $\mathbf{d}^{m} = (u^{m}_{1}, u^{m}_{2}, ..., u^{m}_{t})$ encompasses $t$ utterances between ES agent, which we refer to as the system, and the user. The target response is represented by $\mathbf{r}^{m}$. Let $\mathbf{\phi}^{m} = \{\mathbf{\phi}^{m}_{S}, \mathbf{\phi}^{m}_{U}\}$ represents the set of bidirectional cognitive knowledge (BCK) from $\mathbf{d}^{m}$, where  $\mathbf{\phi}^{m}_{S}$ and $\mathbf{\phi}^{m}_{U}$ denote the  perceptions of system and user, respectively.
 The situational synopsis of the user is represented by $\mathbf{s}$. Our goal is to generate the target response given the set of accumulated $\mathbf{\phi}^{m}$ by estimating the probability distribution $p(\mathbf{r}^{m} | \mathbf{d}^{m}, \mathbf{\phi}^{m}, \mathbf{s})$.
For ease of notation, we exclude the superscript $m$ as we discuss our method on an instance of ES scenario in the following sections.

\subsection{Dynamic Discourse Context Propagation Window} 
To foster timely contextual relevancy distinct to $u_{\psi} \in \mathbf{d}$ where $\psi \in t$ as we construct BCK, we assign a discourse context propagation window $\mathcal{\theta}$ to  $u_{\psi} \in \mathbf{d}$:
    \begin{equation} \label{eq:6}
    \begin{aligned}
    \mathcal{\theta} = \sum_{k=1}^{n} u_{\psi - k}
    \end{aligned}
    \end{equation}
where $n$ is a parameter that determines the local discourse analysis span. To preserve discourse coherence, $n$ should be relatively small in relation to $t$ to foster lexical coherence \cite{halliday2014cohesion}. By introducing $\mathcal{\theta}$, we update the local discourse context scope of BCK synthesis for $u_{\psi} \in \mathbf{d}$. Moreover, since we circumscribe bidirectional cognitive analysis within $\mathcal{\theta}$, we provide traceability to the source of the cognitive knowledge, promoting the interpretability of Mind2 on a knowledge provenance basis.

\subsection{Bidirectional Cognitive Knowledge (BCK) Synthesis}
We approach BCK synthesis with a prompt-based method that bridges discourse analysis with three cognitive theories. We design prompts to execute query expansion, collecting bidirectional cognitive context from $\theta$ to acquire BCK denoted by $\phi$ based on cognitive theories: 1) bidirectional Theory-of-Mind (BTM / $\phi^{BTM}$), 2) BTM-based psychological expected utility ($\phi^{PEU}$), and 3) BTM-based cognitive rationality ($\phi^{BCR}$). Each component of BCK consists of two aspects, one from the system's perspective, $\phi_{S}$, and another from the user's viewpoint, $\phi_{U}$. 
By utilizing query expansion in constructing BCK for context modeling, we directly locate salient information from ES dialogues, preserving context in their original form, which bypasses potential misinterpretation of contextual information from intermediate modification. Our method can be realized with cost-effective LLM, such as GPT-3.5 Turbo. Our prompt template is abstracted as below:
\begin{quote}
    \rule{\linewidth}{1.5pt}
    \textbf{\textcolor{black}{[ROLE]} }
    \textcolor{teal}{Cognitive psychologists specializing in }\textbf{\textcolor{blue}{cognitive theories}}\\
    \textbf{[TASK] }\\
    \textbf{\textit{Step 1:}} \textcolor{teal}{Define} \textbf{\textcolor{blue}{$\theta$}} \\
    \textbf{\textit{Step 2:}} \textcolor{teal}{Perform}
    \textbf{\textcolor{blue}{[BCK synthesis subtask]}} \\
    \textbf{\textit{Input:}} 
    \textbf{\textcolor{blue}{[Input]}}\\
    \textbf{\textit{Output:}} \textbf{\textcolor{blue}{[Output]}}\\
    \textcolor{teal}{{Accumulate extracted terms that you prioritize to derive the \textbf{\textcolor{blue}{[BCK synthesis subtask]}}. If no term is identified, output none.}}
    \rule{\linewidth}{1.5pt}
    
\end{quote}

Next, we will discuss the subtask of BCK synthesis that consists of three components of bidirectional cognitive knowledge synthesis.

\definecolor{x1}{rgb}{0.68, 1, 0.6}
\definecolor{x}{rgb}{0.78, 1, 0.45}
\definecolor{y}{rgb}{0.88, 1, 0.55}
\definecolor{z}{rgb}{0.98, 1, 0.8}
\newcommand\Tstrut{\rule{0pt}{2.6ex}} 
\newcommand\Bstrut{\rule[-0.9ex]{0pt}{0pt}}

\begin{table*}
 
\centering

\caption{Performance of Mind2 trained with 10\% of training data (Mind2-10\%) compared against the performance of baseline models reproduced or reported in \cite{kemi} \cite{bao-etal-2024-multi-stream}. The top two performances of each evaluation metric are highlighted in green with darker gradient indicating better performance.}

\label{tab:main}
\begin{tabular}{p{21mm}p{13mm}p{13mm}p{13mm}p{13mm}p{13mm}p{13mm}p{13mm}}
\Xhline{4\arrayrulewidth}
\Tstrut
\textbf{Model} & \textbf{F1} \trianbox1{cblue} & \textbf{PPL} \uptrianbox1{cblue} & \textbf{B-2} \trianbox1{cblue} & \textbf{B-4} \trianbox1{cblue} & \textbf{D-1} \trianbox1{cblue} & \textbf{D-2} \trianbox1{cblue} & \textbf{R-L} \trianbox1{cblue} \\
\midrule
\textbf{Transformer} & - & 81.55 & 5.66 & 1.31 & 1.29 & 6.91 & 14.68 \\
\textbf{MoEL} & - & 62.93 & 5.02 & 1.14 & 2.33 & 15.26 & 14.21 \\
\textbf{MIME} & - & 43.27 & 4.82 & 1.03 & 2.11 & 10.94 & 14.83 \\
\midrule
\textbf{BlenderBot} & 18.20 & 16.48 & 5.93 & 1.62 & 3.76 & 21.22 & 15.34 \\ 
\textbf{BlenderBot-Joint} & 19.23 &  16.15 & 5.52 & 1.29 & 3.60 & \cellcolor{x1} 21.88 & 15.51 \\
\textbf{GLHG} & - & \cellcolor{y} 15.67 & 7.57 & 2.13 & 3.50 & \cellcolor{y} 21.61 & 16.37 \\
\textbf{MISC} & 19.89 & 16.08 & 7.62 & 2.19 & \cellcolor{x1} 4.41 & 19.71 & 16.40 \\
\textbf{KEMI} & \cellcolor{y} 24.66 & 15.92 & \cellcolor{y} 8.31 & \cellcolor{y} 2.51 & - & - & \cellcolor{y} 17.05 \\
\midrule
\textbf{Mind2}\textit{-10\%} & \cellcolor{x1} \textbf{25.16} & \cellcolor{x1} \textbf{11.66} & \cellcolor{x1} \textbf{13.35} & \cellcolor{x1} \textbf{4.87} & \cellcolor{y} \textbf{3.80} & \textbf{19.64} & \cellcolor{x1} \textbf{20.91} \\
\Xhline{4\arrayrulewidth}
\Tstrut
\end{tabular}
\vspace{-5mm}
\end{table*}
\subsubsection{\textbf{Bidirectional Theory-of-Mind (BTM)}}
Theory-of-Mind (ToM) highlights humans' ability to infer the cognitive state of others, ranging from beliefs to emotions \cite{BARONCOHEN198537}. We adopt the concept of ToM as the basis for fulfilling the speaker's general cognitive understanding of the other speaker. We apply ToM analysis bidirectionally between the system and the user. To determine $\phi^{BTM}_{S}$, our method locates terms from $\theta$ that the system prioritizes in developing its ToM about the user. Likewise, $\phi^{BTM}_{U}$ consists of extracted terms from $\theta$ that the user prioritizes in developing its ToM about the system.

\subsubsection{\textbf{BTM-based Psychological Expected Utility (PEU)}}
We also tackle the bidirectional cognitive analysis of ES from a neuroeconomics premise. Namely, we employ the psychological expected utility (PEU) theory in understanding the ES context. Under PEU, the decision-making process of humans involves utility calculations based on the perceived desirability of potential actions \cite{glimcher2005physiological}. We integrate PEU in our method by accumulating terms in $\theta$ that adhere to the diverging expectations of the system and the user in deriving their PEU. To acquire $\phi^{PEU}_{S}$, we prompt the LLM to select terms from $\theta$ that comply with the system’s expectation to improve the user’s mental state. In contrast, $\phi^{PEU}_{U}$ comprises extracted terms from $\theta$ that comply with the user’s expectation to gain assistance from the system. 

\subsubsection{\textbf{BTM-based Cognitive Rationality (BCR)}}
We further approach bidirectional cognitive analysis from the cognitive rationality perspective. Cognitive rationality explains decision-making based on cognitive confidence optimization, in which an actor chooses the solution that leads to their optimal fulfillment of psychological confidence in response to stimuli \cite{ezhkova2004principles}. We analyze the psychological confidence of the system and the user based on the rationality manifestation. We prompt the LLM to extract terms from $\theta$ that indicate the system rationale for understanding the user's immediate cognitive, emotional, and behavioral states that reflect cognitive rationality to acquire $\phi^{BCR}_{U}$. Similarly, $\phi^{BCR}_{U}$ consisted of salient terms from $\theta$ indicating the user's cognitive rationality resulting in the system's immediate cognitive, emotional, and behavioral states within $\theta$.

\subsection{Bidirectional ES Dialogue Generation}
Following standard practice, we adopt the pre-trained language model BlenderBot \cite{roller} as the model backbone and reformulate the task as a Seq2Seq problem. We first process dialogue history $\mathbf{d}$ and bidirectional cognitive knowledge $\mathbf{\phi}_{S}$ and $\mathbf{\phi}_{U}$ into distributed representations. We encode $\mathbf{d}$ as follows:
    \begin{equation} \label{eq:1}
    \begin{aligned}
    \alpha &= [syp], \mathbf{s}, [\omega_{S}], d_{1}, [\omega_{U}], d_{2}, ..., [\omega_{S}], d_{t}
    \end{aligned}
    \end{equation}
where $[syp]$, $[\omega_{S}]$, and $[\omega_{U}]$ are inserted tokens to differentiate the user's situational context and the roles of speakers, respectively. 

Then, we encapsulate acquired bidirectional cognitive knowledge as triplet sets corresponding to their respective $u_{\psi}$. For instance, we incorporate $\phi_{S}$ for $d_{1}$ as follows:
    \begin{equation} \label{eq:2}
    \begin{aligned}
    \phi_{S} &= [mind], \phi^{BTM}_{S_{1}}, [util], \phi^{PEU}_{S_{1}}, [prnt], \phi^{BCR}_{S_{1}}
    \end{aligned}
    \end{equation}
where [mind], [util], and [prnt] are specialized tokens that correspond to the three components of cognitive knowledge of Mind2 appended by their respective content, $\phi^{BTM}_{S_{1}}$, $\phi^{PEU}_{S_{1}}$, and $\phi^{BCR}_{S_{1}}$.

\definecolor{x1}{rgb}{0.68, 1, 0.6}
\definecolor{x}{rgb}{0.78, 1, 0.45}
\definecolor{y}{rgb}{0.88, 1, 0.55}
\definecolor{z}{rgb}{0.98, 1, 0.8}

\begin{table*}
 
\centering

\caption{Performance of Mind2 given 10\%, 25\%, 50\%, 75\%, and 100\% of training data. The relative change of Mind2-25\%, Mind-50\%, Mind-75\%, and Mind2-100\% against Mind2-10\% are also reported. }

\label{tab:data}
\begin{tabular}{p{9mm}p{27mm}p{10mm}p{10mm}p{10mm}p{10mm}p{10mm}p{10mm}p{9mm}}
\Xhline{4\arrayrulewidth}
\Tstrut
\textbf{Model} & \textbf{\% of Training Data, Relative Change (RC)} & \textbf{F1} \trianbox1{cblue} & \textbf{PPL} \uptrianbox1{cblue} & \textbf{B-2} \trianbox1{cblue} & \textbf{B-4} \trianbox1{cblue} & \textbf{D-1} \trianbox1{cblue} & \textbf{D-2} \trianbox1{cblue} & \textbf{R-L} \trianbox1{cblue} \\
\midrule
\Bstrut
\textbf{Mind2} & \textbf{\textit{10\%}} & 25.16 & 11.66 & 13.35 & 4.87 & 3.80 & 19.64 & 20.91\\

\cline{2-9}
\Tstrut
 & \textbf{\textit{25\%}} & 26.46 & 10.56 & 13.48 & 5.21 & 3.89 & 19.76 & 21.70 \\
 & \textit{10\% $\xrightarrow{}$ 25\% RC} & $\uparrow$\textcolor{teal}{5.2\%} & $\uparrow$\textcolor{teal}{9.5\%} & $\uparrow$\textcolor{teal}{1.0\%} & $\uparrow$\textcolor{teal}{7.0\%} & $\uparrow$\textcolor{teal}{2.4\%} & $\uparrow$\textcolor{teal}{0.6\%} & $\uparrow$\textcolor{teal}{3.8\%}\\

\cline{2-9}
\Tstrut
 & \textbf{\textit{50\%}} & 29.52 & 9.69 & 16.68 & 6.68 & 4.26 & 22.31 & 24.21 \\
 & \textit{10\% $\xrightarrow{}$ 55\% RC} & $\uparrow$\textcolor{teal}{17.3\%} & $\uparrow$\textcolor{teal}{17.0\%} & $\uparrow$\textcolor{teal}{24.9\%} & $\uparrow$\textcolor{teal}{37.2\%} & $\uparrow$\textcolor{teal}{12.1\%} & $\uparrow$\textcolor{teal}{13.6\%} & $\uparrow$\textcolor{teal}{15.8\%}\\
 
\cline{2-9} 
\Tstrut
 & \textbf{\textit{75\%}} & 31.53 & 9.15 & 18.86 & 8.07 & 4.50 & 23.39 & 26.26 \\
 & \textit{10\% $\xrightarrow{}$ 75\% RC} & $\uparrow$\textcolor{teal}{25.3\%} & $\uparrow$\textcolor{teal}{21.6\%} & $\uparrow$\textcolor{teal}{41.3\%} & $\uparrow$\textcolor{teal}{65.7\%} & $\uparrow$\textcolor{teal}{18.4\%} & $\uparrow$\textcolor{teal}{19.1\%} & $\uparrow$\textcolor{teal}{25.6\%}\\
 
\cline{2-9}
\Tstrut
 & \textbf{\textit{100\%}} & \textbf{32.96} & \textbf{8.76} & \textbf{20.11} & \textbf{8.92} & \textbf{4.68} & \textbf{23.51} & \textbf{27.76} \\
 & \textit{10\% $\xrightarrow{}$ 100\% RC} & $\uparrow$\textcolor{teal}{31.0\%} & $\uparrow$\textcolor{teal}{24.9\%} & $\uparrow$\textcolor{teal}{50.6\%} & $\uparrow$\textcolor{teal}{83.2\%} & $\uparrow$\textcolor{teal}{23.2\%} & $\uparrow$\textcolor{teal}{19.7\%} & $\uparrow$\textcolor{teal}{32.8\%}\\
 
\Xhline{4\arrayrulewidth}
\Tstrut 

\end{tabular}
\vspace{-7mm}
\end{table*}

We concatenate $\alpha$ and $\phi_{S}$ and incorporate the start token $[CLS]$ at the beginning of each dialogue context. Together, the linearized input of Seq2Seq learning is
    \begin{equation} \label{eq:3}
    \begin{aligned}
    \Omega = [CLS], \alpha, [cog], \phi_{S_{1}}, \phi_{U_{2}}, ..., \phi_{S_{t}}
    \end{aligned}
    \end{equation}
where [cog] denotes the specialized token for incorporating cognitive knowledge. The linearized output is represented as:
    \begin{equation} \label{eq:4}
    \begin{aligned}
    Y = [str], y, [rsp], r
    \end{aligned}
    \end{equation}
where [str] and [rsp] are inserted tokens for the ES strategy and the generated response. Next, we utilize the linearized input $\Omega$ and generated result $Y$ to optimize the model by maximizing the negative log-likelihood:
    \begin{equation} \label{eq:5}
    \begin{aligned}
    \mathcal{L}_{nll} = -\frac{1}{M}\sum^{M}_{m=1} log P(Y | Y_{<m}, \Omega).
    \end{aligned}
    \end{equation}

\section{Experiments}
\subsection{Experimental Settings}
\subsubsection{\textbf{Evaluation Dataset}}
To evaluate our method, we use the public ESConv dataset \cite{liu-etal-2021-towards}, a collection of 1,300 crowd-sourced emotion support conversations between supporters and support seekers. In all support scenarios, support seekers first take the initiative in the support session by disclosing their problem associated with one of the five categorical distress stimuli -- ongoing depression, job crisis, break up with a partner, problems with friends, and academic pressure. Supporters then employ strategic support techniques (e.g., questions, restatement or paraphrasing, reflection of feelings, self-disclosure, affirmation and reassurance, providing suggestions, information, and others) in their responses to reduce support seekers' emotional distress. 

Instead of adopting the standard 70/15/15 split of training, validation, and testing sets, we utilize just 10\% of the original training data. In contrast, the state-of-the-art baselines we use for comparison use 100\% of their training data. This allows us to assess our model's effectiveness to examine our method's performance in generating effective dialogues given limited in-domain training data. In an actual deployment, a model must stay robust in performing strategic dialogue generation, adapting to the constant inflow of new ES scenarios not necessarily represented in the model's training data.

\subsubsection{\textbf{Implementation Details}}
We implement all models using PyTorch \cite{paszke2019pytorch} and adopt the small version of BlenderBot \cite{roller}. 
During the training stage, we set the training epochs and $\theta$ to five. The warm-up step is set to 100 with a learning rate of 3e-5. The maximum sequence and decoder input lengths are set to 256 and 40, respectively. The training and evaluation batch sizes are both 16. For inference, we adopt decoding algorithms as in Deng et al. \cite{kemi}, setting the top-$k$ and top-$p$ sampling with $p$=0.3 and $k$=30, respectively. The temperature $\tau$ and repetition penalty are set to 0.7 and $1$, respectively. All experiments are conducted on NVIDIA Quadro RTX 4000.

\subsubsection{\textbf{Evaluation Baselines and Metrics}}
We compare the performance of Mind2 against the Transformers-based and BlenderBot-based baselines. The Transformers-based baselines consist Transformers \cite{vaswani2017attention}, MoEL \cite{lin-etal-2019-moel}, and MIME \cite{majumder-etal-2020-mime}. BlenderBot-based methods are BlenderBot \cite{roller}, BlenderBot-Joint \cite{liu-etal-2021-towards}, GLHG \cite{Peng2022ControlGU}, MISC \cite{tu-etal-2022-misc}, and KEMI \cite{kemi}.  
We evaluate models on three dimensions. First, we use macro F1 as the metric for ES strategy prediction. Following automatic evaluation convention \cite{liu-etal-2021-towards} \cite{tu-etal-2022-misc} \cite{kemi}, we evaluate ES dialogue generation from semantic and lexical scopes by adopting perplexity (PPL), BLEU-2 (B-2), BLEU-4 (B-4), and ROUGE-L (R-L). To assess the dialogue generation diversity of models, we record Distinct-1 (D-1) and Distinct-2 (D-2) values, which measure the unique $n$-grams ratios in generated dialogues.

\subsection{Main Evaluation Results}
Table \ref{tab:main} displays the performance of Mind2 against baseline models. Although trained with only 10\% of the available training data, Mind2 (Mind2-10\%) achieves the best performance in ES strategy prediction on all evaluation metrics except D-$n$. The D-$1$ score of Mind-10\% places second, and the difference between the D-$2$ scores of BlenderBot-Joint and Mind2-10\% is 2.24. The B-$n$ metrics show the most dramatic performance enhancement. Compared to the best-performing baseline, the B-2 score of Mind2 exhibits a 60\% performance increase (8.31 $\xrightarrow{}$ 13.35). For B-4 score, Mind2 outperforms the best baseline by 94\% in terms of relative change (2.51 $\xrightarrow{}$ 4.87). The PPL and R-L scores are enhanced by 3.99 and 3.86 points, respectively. We attribute the performance improvement to Mind2's incorporation of context through triplet sets of ToM-based knowledge tailored to each support scenario.

\begin{figure*}
    \centering
    \includegraphics[width=1\linewidth]{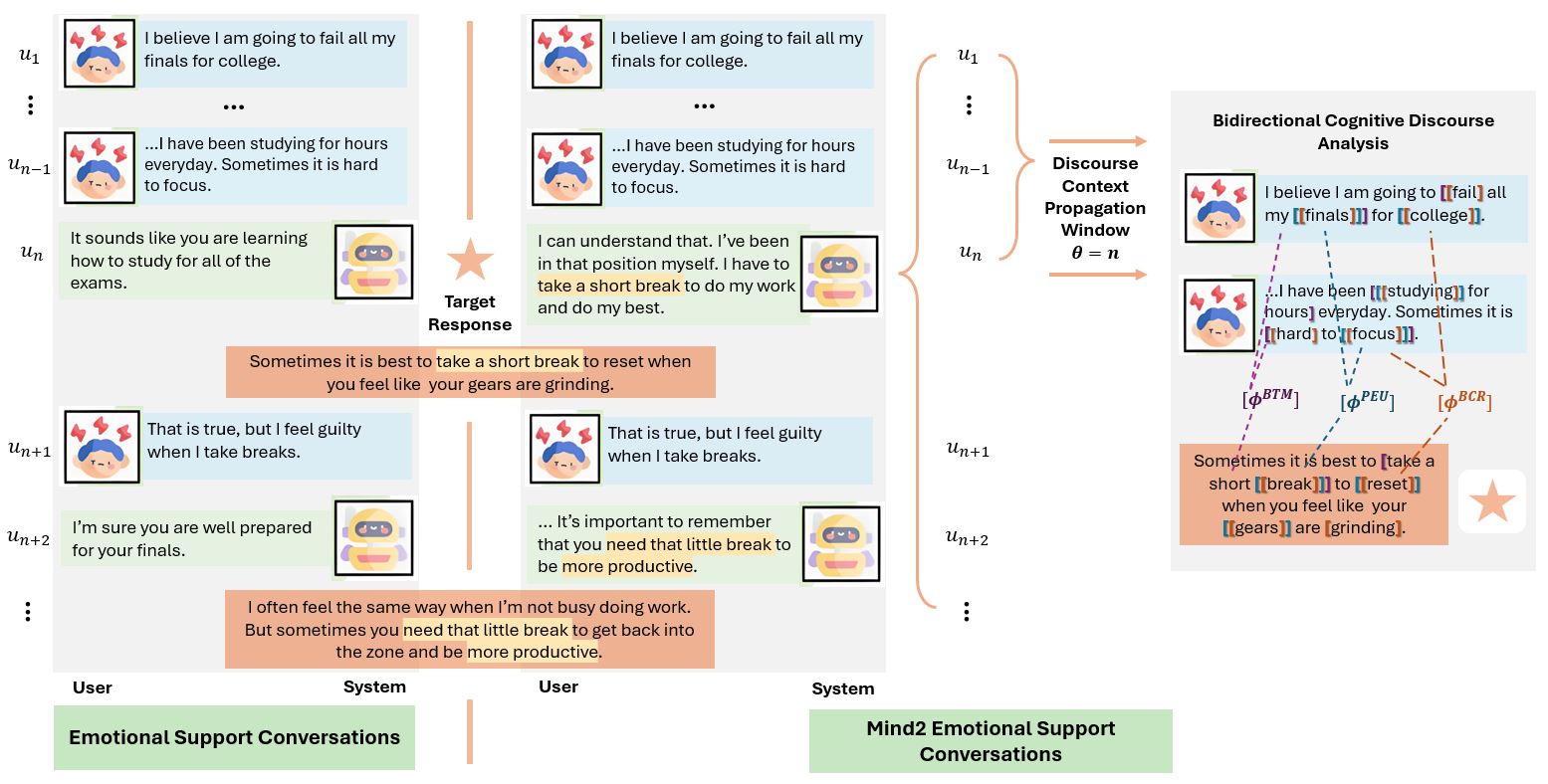}
    \caption{An example of Mind2 ES dialogue generation. In contrast to the baseline, Mind2 model enhanced with three BCK components produces ES dialogues that are contextually more congruent with the target responses. }
    \label{fig:v7}
  
\end{figure*}

\definecolor{x1}{rgb}{0.68, 1, 0.6}
\definecolor{x}{rgb}{0.78, 1, 0.45}
\definecolor{y}{rgb}{0.88, 1, 0.55}
\definecolor{z}{rgb}{0.98, 1, 0.8}

\begin{table*}
 
\centering

\caption{Ablation study on the effect of each bidirectional cognitive knowledge component ($\phi^{BTM}$, $\phi^{PEU}$, and $\phi^{BCR}$)  of Mind2. The top two performances of each evaluation metric are highlighted in green with darker gradient indicating better performance.}

\label{tab:aba}
\begin{tabular}{p{9mm}p{24mm}p{12mm}p{12mm}p{12mm}p{12mm}p{12mm}p{12mm}p{10mm}}
\Xhline{4\arrayrulewidth}
\Tstrut
\textbf{Model} & \textbf{Knowledge} & \textbf{F1} \trianbox1{cblue} & \textbf{PPL} \uptrianbox1{cblue} & \textbf{B-2} \trianbox1{cblue} & \textbf{B-4} \trianbox1{cblue} & \textbf{D-1} \trianbox1{cblue} & \textbf{D-2} \trianbox1{cblue} & \textbf{R-L} \trianbox1{cblue} \\
\midrule
\textbf{Mind2}& \textit{w/o $\phi^{BTM}, \phi^{PEU}$} & 24.27 & 12.34 & 12.12 & 4.58 & 3.85 & 18.35 & 20.31 \\
& \textit{w/o $\phi^{BTM}, \phi^{BCR}$} & 24.36 & 12.20 & 12.17 & 4.66 & 4.21 & 20.09 & 20.15 \\
& \textit{w/o $\phi^{PEU}, \phi^{BCR}$} &  27.90 &  10.45 & 15.56 & 5.98 & 4.52 &  22.47 & 23.23 \\
& \textit{w/o $\phi^{BTM}$} & 28.82 & 10.51 &  16.58 &  7.23 &  4.33 & 21.98 &  24.2 \\
& \textit{w/o $\phi^{PEU}$} & \cellcolor{y} 31.05 & 9.42 & \cellcolor{y} 18.36 & \cellcolor{y} 7.69 & \cellcolor{x1} 4.68 & \cellcolor{x1} 23.89 & \cellcolor{y} 26.02 \\
& \textit{w/o $\phi^{BCR}$} & 25.53 & \cellcolor{y} 9.41 & 18.11 & 7.65 & \cellcolor{y} 4.59 &  23.02 & 25.53 \\
\midrule
\textbf{Mind2} &  \textit{w/} $\phi^{BTM}, \phi^{PEU}$ \textit{, and}  $\phi^{BCR}$ & \cellcolor{x1} \textbf{32.96} &  \cellcolor{x1} \textbf{8.76} &  \cellcolor{x1} \textbf{20.11} &  \cellcolor{x1} \textbf{8.92} & \cellcolor{x1}  \textbf{4.68} &  \cellcolor{y} \textbf{23.51} & \cellcolor{x1} \textbf{27.76} \\

\Xhline{4\arrayrulewidth}
\Tstrut 

\end{tabular}
\vspace{-7mm}
\end{table*}

\subsection{Mind2 Performance vs. Data Utilization} Table \ref{tab:data} displays the performance of Mind2 given varying percentages of training data. We also report the relative performance enhancement concerning the Mind2 model trained with 10\% training data (Mind2-10\%) as adopted when compared against baseline ES systems. As shown, though Mind2-10\% shows competitive performance against baseline models, increasing the amount of training data, which provides the model with more variety of ES scenarios, further improves the performance of Mind2 on all evaluation metrics. Such performance gain due to the increase in training data is especially apparent on the B-$n$ metrics, as reflected through the 50.6\% and 83.2\% relative change in B-$2$ and B-$4$ scores, respectively, when the percentage of training data scaled from 10\% to 100\%.
Overall, relative change exhibits the greatest incremental enhancement across metrics when we increase the training data percentage from 25\% to 50\%. By utilizing more training data, Mind2-75\% and Mind2-50\% surpass the performance of baselines in D-$1$ and D-$2$ in Table 1, respectively. The Mind2 model utilizing the entire training dataset achieves the best performance against all evaluation metrics. 

\subsection{Analysis on Bidirectional Cognitive Knowledge} 
We examine the effect of three components of BCK -- $\phi^{BTM}$, $\phi^{PEU}$, and $\phi^{BCR}$ -- on the performance of Mind2. Figure \ref{fig:v7} illustrates an ES scenario with ES dialogues generated with the baseline Mind2 model that does not incorporate any BCK and the Mind2 model enhanced with three BCK components. The shared terms highlighted in the light green of BCK-equipped generate dialogues with their respective target responses support that integrating BCK improves contextual relevancy in ES dialogue generation. Quantitatively, we conducted an ablation study (Table \ref{tab:aba}) and show that Mind2 demonstrates the best performance on all evaluation metrics besides the D-$2$ score when all three components of BCK are utilized, validating the respective qualitative dialogue generation performance in Figure \ref{fig:v7}.

On average, the performance of Mind2-\textit{w/o} $\phi^{PEU}$ ranks second, followed by the performance of Mind2-\textit{w/o} $\phi^{BCR}$. We also observe that Mind2 models without $\phi^{BTM}$ exhibit weaker performance than their counterparts utilizing the same count of BCK component. To interpret the varying effectiveness of each BCK component in ES dialogue generation, we analyze the semantic relevance of terms to their respective cognitive theory. Analysis reveals 
$\phi^{BTM}$ contains the highest percentage (73.6\%) of semantically significant terms (e.g., excluding terms such as \textit{none}), followed by $\phi^{BCR}$ (65.3\%) and $\phi^{PEU}$ (61.1\%). Thus, we infer that the relative weight of $\phi^{BTM}$, $\phi^{BCR}$, and $\phi^{PEU}$ in uplifting the performance of Mind2 correlates with their varying contextual relevancy to their respective cognitive theory. Hence, due to its most informative semantic representation reflecting BTM, introducing  $\phi^{BTM}$ to Mind2 results in the highest performance gain in Mind2's ES dialogue generation.

\section{Conclusions and Future Work}
We introduce Mind-to-Mind (Mind2), an ES framework that models contexts for ES dialogue generation with bidirectional cognitive discourse analysis. We incorporate Theory-of-Mind, psychological expected utility, and cognitive rationality while constructing BCK with a prompt-based query expansion method for ES dialogue generation. Experimental results support the competitive performance of Mind2 against state-of-the-art ES systems with 10\% of the training data and that Mind2 surpasses all baselines as we increase the percentage of training data. Further analysis supports the effectiveness of three BCK components and provides insights on their relative significance in enhancing the ES dialogue generation performance. 

In future work, we want to experiment with the integration of interpretability in ES context modeling with a greater extent of context adaptability. Given the limited in-domain dataset, we consider model adaptability from two bases. First, we want to enhance our control of the context adaptability on a local level. In Mind2, $\theta$ controls the source of BCK, which treats the defined discourse span as a parameter. We want to explore techniques where the model learns to set the local discourse span for each utterance based on significant contextual shifts. We also consider context adaptability from a domain adaptation angle. We will design methods for ES systems to effectively bridge potential knowledge gaps in providing user-centric support by integrating out-of-domain information, such as user personality characteristics. 

\section{Acknowledgements}
We thank all reviewers for their constructive feedback. This work is supported by the University of Arkansas's Honors College Research Grant and NSF 1946391 RII Track-1: Data Analytics that are Robust and Trusted (DART): From Smart Curation to Socially Aware Decision Making.

\bibliographystyle{ieeetr}
\bibliography{WI237}

\end{document}